# Research on Flight Accidents Prediction based Back Propagation Neural Network

Haoxing Liu[1], Fangzhou Shen[2], Haoshen Qin[3] and, Fanru Gao[4(*)]

[1] Flight Department, Shanghai Jixiang Airlines Co., Ltd., Shanghai, China;
[2] Department of Mathematics and Statistics, San Jose State University , San Jose, USA;
[3] Department of Computer & Information Science & Engineering, Herbert Wertheim College of Engineering, University of Florida, Gainesville, 32611, USA;
[4] Department of Mechanical and Aerospace Engineering, Case Western Reserve University, Cleveland, OH, 44106, USA.
\*fxg149@case.edu

**Abstract.** With the rapid development of civil aviation and the significant improvement of people's living standards, taking an air plane has become a common and efficient way of travel. However, due to the flight characteristics of the aircraft and the sophistication of the fuselage structure, flight delays and flight accidents occur from time to time. In addition, the life risk factor brought by aircraft after an accident is also the highest among all means of transportation. In this work, a model based on back-propagation neural network was used to predict flight accidents. By collecting historical flight data, including a variety of factors such as meteorological conditions, aircraft technical condition, and pilot experience, we trained a backpropagation neural network model to identify potential accident risks. In the model design, a multi-layer perceptron structure is used to optimize the network performance by adjusting the number of hidden layer nodes and the learning rate. Experimental analysis shows that the model can effectively predict flight accidents with high accuracy and reliability.

## 1  Introduction

With the continuous increase and expansion of routes, the gradual increase in flight mileage and the continuous increase in the number of flights, most airlines generate a large amount of flight data in their daily aircraft operations. These data need to be recorded and saved for later statistics and analysis [1]. The main data can be divided into three categories: flight hour data, attachment cumulative time data, and fault replacement data. These three types of data are the most authoritative indicators of a flight's safety and reliability. If we can sort out and analyze the fault data generated in aircraft operation and establish a reasonable prediction model to predict the potential failure risk, we can prevent it in advance and solve it in time, so as to significantly improve the safety and reliability of the flight and ensure the safety of passengers and crew [2]. Therefore, it is feasible and important to use an accurate neural network prediction model to predict the probability of aircraft failure.

In the aviation industry, flight safety is always one of the most important concerns. As air traffic continues to increase, so does the potential risk of flight accidents, which is a huge challenge for airlines, safety regulators, and the public. Although the rate of accidents is relatively low, the consequences of each accident can be extremely serious, including large-scale casualties and economic losses [3]. Therefore, improving the effectiveness of accident prevention measures, especially by identifying risks and taking preventive measures in advance, is key to improving overall aviation safety.

With the rapid development of the modern civil aviation industry and the frequent occurrence of various civil aviation incidents, more and more scholars have begun to pay attention to the aviation industry. These studies cover a wide range of aspects, from aircraft manufacturing and aircraft structure to the flight environment and the failure hazards of aircraft [4]. Early civil aircraft were mainly produced by two major manufacturers, Boeing and Airbus, and the repair of these aircraft followed standardized maintenance standards. Different engine models and models have significant differences in the occurrence of failures. Therefore, in the study of aircraft failures, in most cases, it is a data analysis study that is customized for a specific aircraft type or equipment type of a particular airline [5]. Among the many predictive technologies, AI offers an efficient way to process and analyze large amounts of flight data. In particular, backpropagation neural network-based methods have proven to be very effective in solving many complex nonlinear problems [6].

## 2 Related Work

Compared with other industries, the research in the field of civil aviation is still a relatively niche research field due to its industry specificity and the sensitivity of aircraft equipment. Much of the research is usually limited to a specific type of aircraft engine or component. There are relatively few cases where a comprehensive study of an airline's day-to-day operational failures has been studied [7]. However, it is relatively mature in the research and application of big data and neural networks. Through the rapid processing technology of big data, the huge operation data can be quickly classified, and the qualified sample data can be screened out as input data, which provides new opportunities for research in the field of civil aviation [8].

Initially, the study by Yiru [9] explores the impact of aircraft pillar systems on aircraft safety. By analyzing the multi-dimensional data of the aircraft pillar system and conducting multiple simulated crash experiments, the research team conducted a detailed analysis of the experimental data. Subsequently, researcher Wu Jiang's [10] utilized the Learning vector quantization (LVQ) neural network to model and predict the failure of aircraft shock absorbers. In this study, various parameters in the maintenance information were comprehensively analyzed, including the service time, expiration date and flight time of the spare parts as inputs, and the failure prediction model of LVQ neural network was established with oil leakage as the output. The model is trained on historical maintenance information, and then the current maintenance information is fed into the trained prediction model to achieve the expected prediction effect. Additionally, Zhou [11] proposed and simulated the Principal components analysis and Back-propagation neural network (PCA-BP) model to address the challenges of product quality prediction in modern industry.

## 3 Methodologies

### 3.1 Back-propagation Neural Network

The back-propagation neural network is a multi-layer feedforward artificial neural network. The architecture consists of an input layer, one or more hidden layers, and an output layer. The input layer receives a variety of predictors such as weather conditions, mechanical data, pilot flight hours, and flight history. The hidden layer enables the model to learn complex patterns from the data, and the output layer produces predictions about whether a flight accident will occur. Output $y$ of each neuron is defined by the application of the activation function to input-weighted sum shown as Equation 1.

$$y = f\left(\sum_{i=1}^{N} w_i \cdot x_i + b\right) \#(1)$$

Where $x_i$ is the input of the neuron, $w_i$ is the weight, $b$ is the bias, and $f(\cdot)$ is the tanh activation function. Data is collected from historical flight records, both those that have had accidents and those that have not. This data is then normalized to ensure that all input features contribute equally to the model training process. Normalization typically involves scaling the data to a mean of zero and a standard deviation of one.

In forward propagation, input data is passed from the input layer through the hidden layer to the output layer. Each neuron passes the result forward after weighting its input and applying an activation function. Use backpropagation to update the weights and biases of neurons. Chief is calculated by the loss function, which measures the difference between the predicted output and the actual output. Mean squared error is commonly used, and its calculation process is shown in Equation 2.

$$MSE = \frac{1}{N}\sum_{i=1}^{N}(y_i - \hat{y}_i)^2 \#(2)$$

Where $\hat{y}_i$ is the predicted output and $y_i$ is the actual output.

### 3.2 Loss Function

For the output layer, we first calculate the partial derivative of the loss function for the output of each output neuron. If the MSE and tanh activation functions are used, assuming that the activation function of the output layer is $f(x)$, then the error gradient for each neuron of the output layer can be expressed as following Equation 3.

$$\delta_k = (y_i - \hat{y}_i) \cdot f'(z_k) \#(3)$$

Where $z_k$ is the input of the output layer neuron and $f'(\cdot)$ is the derivative of the activation function. For hidden layers, the gradient depends not only on the output error of that layer, but also on the gradient of the next layer. By the chain rule, the gradient of the $j$ hidden layer neuron is shown as Equation 4.

$$\delta_j = \sum_k w_{jk} \delta_k \cdot f'(z_j) \#(4)$$

Where $w_{jk}$ is the weight that connects the $j$ hidden neuron and the $k$ output neuron, and $\delta_k$ is the gradient of the $k$ output neuron. Additionally, the gradient of the loss function with respect to each weight and bias in the network is calculated using the calculus chain rule. Adjust weights and biases to minimize losses using optimization algorithms such as gradient descent. In each iteration, the rules for each weight $w_{ij}^{(new)}$ and offset $b_i^{(new)}$ are shown in Equation 5.

$$w_{ij}^{(new)} = w_{ij}^{(old)} - \mu \frac{dMSE}{dw_{ij}}$$
$$b_i^{(new)} = b_i^{(old)} - \mu \frac{dMSE}{db_i} \#(5)$$

Where $\mu$ is the learning rate, a hyperparameter that controls the pace of learning. The gradient $\frac{dMSE}{dw_{ij}}$ and $\frac{dMSE}{db_i}$ represents the effect of weights and biases on losses.

## 4  Experiments

### 4.1  Experimental Setups

The National Transportation Safety Board's (NTSB) Aviation Accident Database provides detailed U.S. civil aviation accident data from 1962 to the present, including detailed circumstances of the accident, aircraft information, cause analysis, and casualties. We collected data from aviation databases, including flight logs, mechanical maintenance, pilot qualifications, and meteorological information. Historical flight data is typically collected from a variety of sources, including sensors on the aircraft, air traffic management systems, weather services, and airline operational records. This data is recorded in a flight data recorder and can include the aircraft's speed, altitude, heading, engine status, flight control system inputs, and environmental conditions, among others. During the pre-processing phase, incomplete, incorrect, or inconsistent data is encountered, which is shown in Figure 1.

| Location | Country | Latitude | Longitude | Airport.Co | Airport.Na | Injury.Seve | Aircraft.Da | Aircraft.Ca | Registratio | Make | Model | Amateur.E | Number.o | Engine.Ty |
|---|---|---|---|---|---|---|---|---|---|---|---|---|---|---|
| Elk, CA | United Sta | 39.12861 | -123.716 | LLR | Little River | Non-Fata | Substantia | Airplane | N7095M | Cessna | 175 | No | 1 | Reciproca |
| OLATHE, | United Sta | 38.84611 | -94.7361 | OJC | Johnson C | Fatal(2) | Destroyed | Airplane | N602TF | Mooney | M20S | No | 1 | Reciproca |
| Fairbanks, | United Sta | 64.66695 | -148.133 | | | N/A | Non-Fata | Substantia | N4667C | Cessna | 170 | No | 1 | |
| GRANBUR | United Sta | 32.36556 | -97.645 | | | N/A | Non-Fata | Substantia | N519RV | Vans | RV 10 | Yes | | |
| Missoula, | United States | | | MSO | | | Unavailab | Substantial | N4476B | Cessna | 170 | No | 1 | |
| LAFAYETT | United Sta | 30.17611 | -92.0075 | LFT | Lafayette | Fatal(5) | Destroyed | Airplane | N42CV | Piper | PA 31T | No | 2 | Turbo Pro |
| Headland, | United Sta | 31.36417 | -85.3125 | 0J6 | Headland | Fatal(1) | Substantia | Helicopter | N663SF | Bell | 407 | No | 1 | Turbo Sha |
| San Rafae | Mexico | | | | | | Unavailab | Destroyed | Airplane | | CESSNA | 208 | | |
| Chandler, | United States | | | CHD | | | Unavailab | Substantial | | N924PA | Piper | PA28 | No | 1 | |
| Lake Hava | United Sta | 34.44 | -114.346 | | | N/A | Non-Fata | Substantial | N5057Z | Cosmos | Phase II | No | | |
| Evansville, | United Sta | 38.09583 | -87.5406 | EVV | Evansville | Fatal(1) | Destroyed | Airplane | N601FL | Piper | PA28 | No | 1 | Reciproca |
| Delaware, | United States | | | DLZ | | | Non-Fata | Substantia | N91WW | Flight Des | CTSW | No | | |
| Beeville, T | United Sta | 28.3675 | -97.7964 | BEA | | | Non-Fata | Substantia | Helicopter | N695AP | Robinson | R22 | No | 1 | |
| Caldwell, I | United Sta | 43.64389 | -116.637 | EUL | CALDWEL | Non-Fata | Substantia | Airplane | N1107C | Piper | PA22 | No | 1 | Reciproca |
| Harrison, ( | United Sta | 39.25917 | -84.7744 | I67 | Cincinnati | Non-Fata | Substantia | Airplane | N5406R | Cessna | 172 | No | 1 | Reciproca |

**Fig. 1.** Illustration of used dataset.

After rigorous data preprocessing, such as cleaning, feature selection, and normalization, a BPNN model was established that included an input layer, multiple hidden layers, and a single output layer. The model was trained and evaluated on the data divided into training sets, validation sets, and test sets, focusing on metrics such as accuracy. Through this method, we are able to build a robust model that can effectively predict flight accidents, which provides scientific data support for aviation safety management. Above Figure 1 shows the collected data from NTSB.

### 4.2 Experimental Analysis

Evaluation metrics include accuracy and confusion matrices, which are measures of the performance of a predicted problem. Initially, accuracy is one of the most commonly used performance metrics and measures the total number of correctly predicted (accidents and non-incidents) by the model as a percentage of the total sample size. Following Figure 2 shows the comparison of accuracy results.

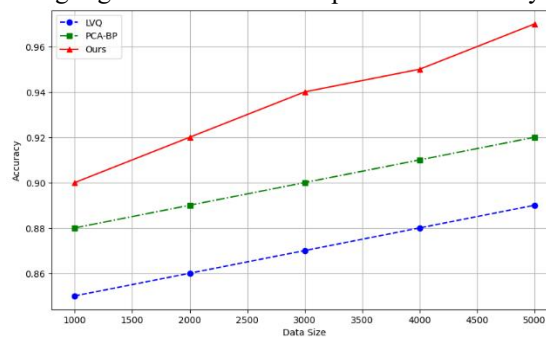

**Fig. 2.** Comparison of Accuracy by Different Methods.

The confusion matrix provides a visual and quantitative way to understand how the model performs on various types of predictions. It shows the intersection of actual and predicted categories, making it possible to visually observe which categories the model performs well and which fall short. Following Figure 3 demonstrates the confusion matrix comparison results.

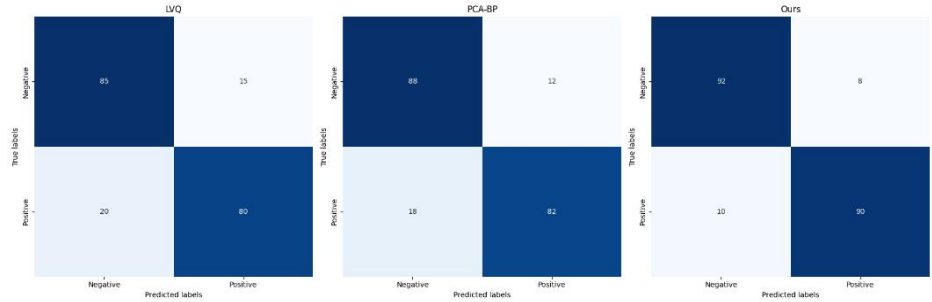

**Fig. 3.** Comparative Analysis of Confusion Matrices for Prediction Models.

## 5  Conclusion

In conclusion, our proposed method introduces a novel adversarial attack method leveraging Generative Adversarial Networks (GANs) to probe the vulnerabilities of image classification systems. Models built by analyzing historical flight data can help predict and reduce flight accidents, optimize flight scheduling and maintenance schedules, and thus improve overall aviation safety. Predictive maintenance models can predict equipment failures and perform maintenance in advance to prevent problems in flight. By generating adversarial samples with imperceptible perturbations, our approach successfully deceives advanced classifiers while maintaining the natural appearance of the images. As for future improvements, exploring more sophisticated GAN architectures and training strategies could enhance the effectiveness of adversarial sample generation, potentially leading to more potent attacks.